\documentclass[runningheads]{llncs}
\usepackage{graphicx}

\usepackage{booktabs}
\usepackage{multirow}
\usepackage{pifont}
\usepackage{colortbl}
\usepackage{amsmath}
\usepackage{amsfonts}

\begin{document}
%

\title{Meply: A Large-scale Dataset and Baseline Evaluations for Metastatic Perirectal Lymph Node Detection and Segmentation}

\titlerunning{Meply: Metastatic Perirectal Lymph Node Segmentation Dataset} 

\author{
Weidong Guo \inst{1,2}
\and
Hantao Zhang \inst{1,2}
\and
Shouhong Wan \inst{1,2} \thanks{\small Corresponding author.}
\and
Bingbing Zou\inst{3,2,4} 
\and
Wanqin Wang\inst{3,2,4} 
\and
Chenyang Qiu \inst{3,2,4}
\and
Jun Li\inst{1} 
\and
Peiquan Jin\inst{1} 
}
\authorrunning{W. Guo et al.}
\institute{School of Computer Science and Technology, University of Science and Technology of China, Hefei, China \\ \and
Institute of Artificial Intelligence, Hefei Comprehensive National Science Center, Hefei, China \\ \and
The First Affiliated Hospital of Anhui Medical University, Hefei, China \\ \and
Anhui Medical University, Hefei, China
}

\maketitle              

\begin{abstract}

Accurate segmentation of metastatic lymph nodes in rectal cancer is crucial for the staging and treatment of rectal cancer. However, existing segmentation approaches face challenges due to the absence of pixel-level annotated datasets tailored for lymph nodes around the rectum. Additionally, metastatic lymph nodes are characterized by their relatively small size, irregular shapes, and lower contrast compared to the background, further complicating the segmentation task. To address these challenges, we present the first large-scale perirectal metastatic lymph node CT image dataset called \textbf{Meply}, which encompasses pixel-level annotations of 269 patients diagnosed with rectal cancer. Furthermore, we introduce a novel lymph-node segmentation model named \textbf{CoSAM}. The CoSAM utilizes sequence-based detection to guide the segmentation of metastatic lymph nodes in rectal cancer, contributing to improved localization performance for the segmentation model. It comprises three key components:  sequence-based detection module, segmentation module, and collaborative convergence unit.  To evaluate the effectiveness of CoSAM, we systematically compare its performance with several popular segmentation methods using the Meply dataset. Our code and dataset will be publicly available at: \url{https://github.com/kanydao/CoSAM}.
\end{abstract}

\section{Introduction}

\begin{figure}[!p]
\begin{center}
\includegraphics[height=1.0\textheight]{preview_final.pdf}
\end{center}
   \caption{An overview of the annotated metastatic perirectal lymph nodes in CT. (a) demonstrates CT sequences of lymph nodes in various sizes. (b) illustrates the  volume distribution of the lymph nodes in our dataset. (b) represents the views from 3 different perspectives and 3D rendering results of the annotations. }
\label{fig:dataset}
\end{figure}

Rectal cancer, the most prevalent type of colorectal cancer, poses an increasingly severe threat to the health and safety of people worldwide \cite{keller2020multidisciplinary}. Precise estimation of rectal lymph node size is crucial for staging patients with rectal cancer, ensuring timely therapeutic management, and evaluating the response to therapy. Specifically, the number of metastatic lymph nodes plays a pivotal role in the pathological examination of rectal cancer for N staging \cite{muthusamy2007optimal}.

Machine learning models for medical image segmentation have shown remarkable progress in recent years \cite{zhang2023customized,zhang2023care}. Nevertheless, to the best of our knowledge, there are currently no tools accessible for comprehensive quantification of metastatic lymph nodes in rectal cancer or for further staging diagnosis. Part of the reason may be the absence of pixel-level ground truth annotations for metastatic lymph nodes in rectal cancer. Primarily, two significant challenges hinder the acquisition of pixel-level annotations for metastatic lymph nodes in rectal cancer. Firstly, metastatic lymph nodes in rectal cancer are closely associated with the staging of rectal cancer. To ensure the dataset's quality, it is crucial to differentiate metastatic lymph nodes based on rectal cancer staging results, necessitating guidance from experienced medical professionals. Second, metastatic lymph nodes frequently have small size, irregular shapes and indistinct borders, making them challenging to identify without medical expertise. Besides, the manual annotation for pixel-level datasets is time-consuming.

While the identification of lymph nodes is challenging, some recent work \cite{LNQ2023,yan2018deeplesion,cardenas2020head,SegRap2023} has made the initial exploration. However, these studies mainly focus on the lymph nodes in their respective body fields(such as the mediastinum, head, and neck). 
Compared to metastatic lymph nodes in other regions of the human body, the anatomical structure of tissues and organs surrounding metastatic lymph nodes in rectal cancer is more complex. Consequently, the identification of metastatic lymph nodes in rectal cancer is more susceptible to interference from neighboring tissues and organs, rendering it a more challenging task. To address this challenge, there is an urgent need for a high-quality, finely annotated dataset of metastatic lymph nodes in rectal cancer. However, to the best of our knowledge, there exists no dataset about the lymph nodes in the area of the rectum. In this study, we collect a large-scale real clinical CT image dataset specifically focused on \textbf{Me}tastatic \textbf{P}erirectal \textbf{Ly}mph node in rectal cancer named \textbf{Meply}, meticulously annotated at the pixel level. An example of CT and annotation from the Meply is illustrated in Fig. \ref{fig:dataset}. For each case in the Meply dataset, a panel of highly experienced doctors with over 20 years of expertise engage in comprehensive discussions. Initially, they focus on identifying the staging of rectal cancer, followed by a precise determination of the location and margins of the metastatic lymph nodes in rectal cancer. In summary, Meply is a large-scale clinical CT dataset exclusively dedicated to metastatic lymph nodes in cases of rectal cancer. 

Compared with natural scenes, medical images tend to be considerably more intricate. \cite{zhang2023decoupling,zhang2024lefusion} A significant gap usually exists between them and natural images. Some mainstream segmentation methods may prove challenging to apply directly to the medical scenes. To tackle this challenge, various segmentation techniques \cite{chen2021transunet,cao2022swin,wang2022uctransnet}, designed explicitly for medical images, have been developed. Nevertheless, these medical image segmentation methods typically target larger organs or substantial lesions, posing difficulties in achieving superior results for more minor, edge-sensitive organs and lesions.

Specifically for metastatic lymph nodes, they are often within the intensity profile of normal soft tissue and have ill-defined borders. As shown in Fig. \ref{fig:dataset}, the perirectal metastatic lymph nodes exhibit low contrast against the background elements, posing challenges in boundary delineation. From a voxel distribution perspective, the majority of perirectal metastatic lymph nodes are composed of fewer than 1600 voxels, indicating a very small volume. Further more, the perirectal metastatic lymph nodes demonstrate a rich diversity in morphology and size. All these factors contribute to the difficulty in localizing perirectal metastatic lymph nodes, directly resulting in reduced accuracy in segmentation tasks.

Previous methods \cite{chen2021transunet,cao2022swin,wang2022uctransnet} can hardly achieve the precise localization of metastatic lymph nodes. Thanks to SAM's promptable paradigm \cite{kirillov2023segment}, box-level prompt information can effectively help the model learn to locate metastatic lymphatic areas. 
However,  recent methods based on SAM \cite{zhang2023customized,zhang2023care} are highly dependent on bounding boxes which utilize more auxiliary information during the test process. These methods can be classified as semi-automatic segmentation. In contrast, we propose a \textbf{Co}llaborative learning framework based on \textbf{SAM} named \textbf{CoSAM} which is no need to use additional box-level prompt information during the reasoning process to achieve fully automatic segmentation.
Moreover, the proposed CoSAM method, by jointly addressing detection and segmentation tasks, effectively decouples to some extent the localization of perirectal metastatic lymph nodes from mask prediction. The model collaboratively optimizes both localization and mask prediction subtasks, thereby better overcoming the negative impact of the challenging localization of perirectal metastatic lymph nodes on segmentation. Additionally, it demonstrates good adaptability to the complex characteristics of perirectal metastatic lymph nodes, including blurred edges and diverse morphological structures.

\begin{figure*}[t]
\begin{center}
\includegraphics[width=\linewidth,scale=1.00]{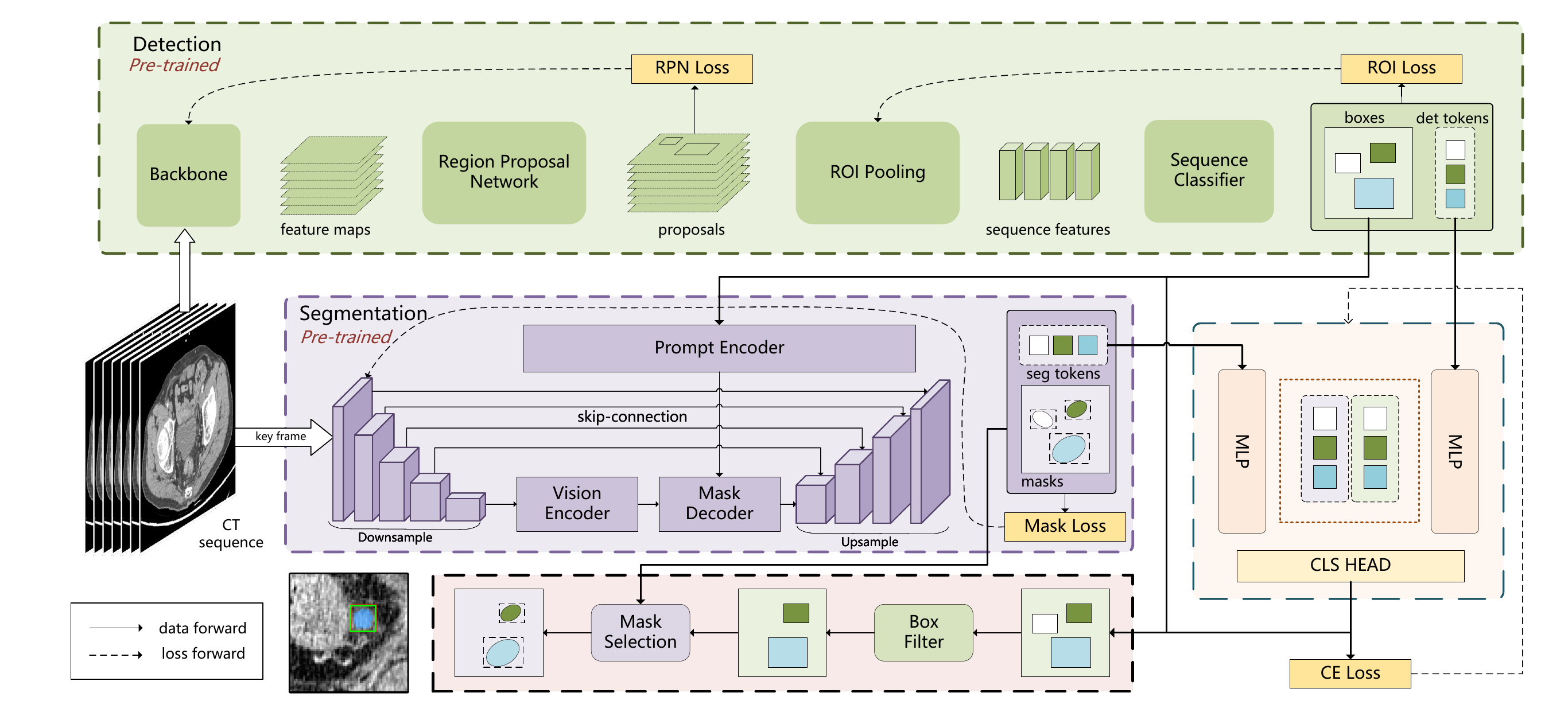}
\end{center}
   \caption{The framework of our proposed CoSAM.}
\label{fig:framework}
\end{figure*}

The contributions of this paper can be summarized as:

(1) As shown in Fig. \ref{fig:framework}, we propose an efficient collaborative learning network framework for segmentation and objecet detection. Through box-level prompt information, detection may aid segmentation to achieve better localization. At the same time, using the segmentation results can better help detect and delete some invalid candidate frames.
(2) We introduce the sequence information between multiple CT frames into the detection of lymph, and use the trajectory of metastatic lymph to better locate it. The introduction of CT's sequence information can enable the detection branch to obtain better detection results.
(3) We construct a large-scale CT image dataset with fine pixel-level annotations for \textbf{Me}tastatic \textbf{P}erirectal \textbf{Ly}mph node detection and segmentation named \textbf{Meply}. Experimental results demonstrate that our proposed CoSAM model obtains substantial improvements compared to existing methods and achieves state-of-the-art on the proposed dataset.

\section{Related Works}

\subsection{Lymph Node Dataset}
Precise estimation of lymph node size holds paramount importance in the staging of cancer patients, guiding initial therapeutic decisions, and evaluating therapy response in longitudinal scans. Nevertheless, this task presents significant challenges, primarily stemming from the low contrast of surrounding structures in Computed Tomography (CT) images and the diverse characteristics of lymph nodes, including their sizes, orientations, shapes, and dispersed positions. The segmentation of all abnormal lymph nodes within a scan offers a promising avenue to assist in diagnosing rectal cancer.

As shown in Table \ref{table:dataset}, there has been some research \cite{LNQ2023,yan2018deeplesion,cardenas2020head,SegRap2023,andrearczyk2021overview,bouget2023mediastinal,roth2014new} focused on the dataset collection of lymph nodes in various anatomical regions, including the mediastinum, head, and neck. However, certain datasets only offer annotations at the bounding-box level, such as DeepLesion \cite{yan2018deeplesion} and 2.5D LN \cite{roth2014new}. Furthermore, there are datasets not explicitly intended for identifying metastatic lymph nodes. Consequently, not all cases within these datasets \cite{yan2018deeplesion} feature annotations for metastatic lymph nodes. In contrast, our Meply dataset is purposefully curated for metastatic lymph nodes in rectal cancer, ensuring that all cases encompass the relevant annotations. Simultaneously, it's worth noting that data in certain public datasets, such as the Mediastinal LN dataset \cite{bouget2023mediastinal}, is not sourced directly from clinical practice. Instead, it has been meticulously curated by aggregating and cleaning data from diverse existing datasets. We carefully chose 269 distinct patients with clearly identifiable metastatic lymph nodes in rectal cancer among individuals diagnosed with rectal cancer from the clinical. Each case in the dataset underwent pixel-level annotation. Given the intricacies of rectal organs and the identification of lymph nodes, this process often necessitated the expertise of seasoned clinicians with extensive surgical experience to render judgments. To the best of our knowledge, our proposed Meply dataset represents the first large-scale CT dataset with finely pixel-level annotations specifically targeting metastatic lymph nodes within the rectal region.

\subsection{SAM in Medical Image Analysis.}
Recently, medical image segmentation has witnessed a significant transformation thanks to the emergence of the Segment Anything Model (SAM), a powerful large-scale vision model \cite{kirillov2023segment}. It provides an excellent interactive segmentation paradigm for prompt-based medical image segmentation. Building upon this paradigm, several research \cite{ma2023segment,wu2023medical,zhang2023care} have been introduced to investigate SAM's potential and its limitations in medical image segmentation. Some of these \cite{wu2023medical,ma2023segment} are predominantly centered on transfer learning techniques. They leverage knowledge acquired from extensive natural image datasets to address specific challenges within the medical domain. Their primary objective is to fine-tune SAM for medical image using techniques like adapter methods. On the other hand, other research efforts have focused on adapting SAM's architecture to better suit the medical domain. For instance, Zhang et al. proposed the U-SAM model \cite{zhang2023care}, specifically tailored to enhance cancer segmentation.

It's worth noting that these SAM-based methods are semi-automatic segmentation approaches, relying on predefined auxiliary prompt information (e.g., bounding boxes or points) during the inference process. 
To alleviate both the SAM model's reliance on additional prompts and the significant negative impact of target localization difficulties on segmentation performance, we decouple the segmentation task into two subprocesses: target localization and mask prediction. Based on this idea, we propose the \textbf{Co}llaborative learning framework based on \textbf{SAM} named \textbf{CoSAM}. This collaborative approach harnesses the power of detection to improve the precision of segmentation, simultaneously employing segmentation outcomes to enhance the detection process and eliminate spurious candidate frames. As an automated segmentation model, our approach no longer relies on additional prompt information. 

\section{The Meply Dataset}

First, lymph nodes are often within the intensity profile of normal soft tissue and have ill-defined borders, which
makes them difficult to identify without medical training. Their presentation across subjects can vary significantly, making it difficult to scale from small datasets to a robust tool. Second, since there are frequently more than one diseased node per case, and manual annotation is time consuming, there are no pre-existing clinical use cases where cases are being
fully annotated. 
Despite such challenges and costs, we present Meply dataset, which is a large scale finely pixel-level annotated dataset of metastatic lymph nodes in rectal cancer.
Researchers and medical practitioners can leverage this dataset to develop and validate segmentation algorithms, pivotal for precise identification and delineation of metastatic perirectal lymph nodes. Such segmentation endeavors are instrumental in treatment planning and disease progression monitoring.

\subsection{Overview}
\begin{table}[t]
\renewcommand\arraystretch{1.5} 
\scriptsize
\begin{center}
\begin{tabular}{c c c c c c c}
\hline\hline
Dataset&Modality&Area&Pixel-level&Number\\
\hline

LNQ2023\cite{LNQ2023}&CT&Mediastinum&\ding{51}&300\\
DeepLesion \cite{yan2018deeplesion}&CT&Body&\ding{55}&4427\\
AAPM-RT-MAC \cite{cardenas2020head}&MRI&Head\&Neck&\ding{51}&55\\
SegRap2023 \cite{SegRap2023}&CT&Nasopharynxk&\ding{51}&200\\
HECKTOR \cite{andrearczyk2021overview}&CT&Head\&Neck&\ding{51}&325\\
Mediastinal LN \cite{bouget2023mediastinal}&CT&Mediastinum&\ding{51}&120\\

\hline
2.5D LN \cite{roth2014new} &MRI&Abdomen&\ding{55}&86\\
\textbf{Meply(ours)}&\textbf{CT}&\textbf{Rectum}&\ding{51}&\textbf{269}\\

\hline\hline
\end{tabular}

\caption{Summary of several publicly available datasets. Modality: Medical data modalities. Area: Body parts covered by the dataset. Pixel-level: Whether the dataset contains pixel-level annotations. Number: the number of CT included in the dataset. }\label{table:dataset}
\end{center}

\end{table}

The \textbf{Me}tastatic \textbf{P}erirectal \textbf{Ly}mph node dataset (Meply), encompassing 269 enhanced computed tomography (CT) scans with a voxel resolution of 0.625mm, is tailored for the specific task of lymph node segmentation. Annotating each scan meticulously, Meply offers invaluable data facilitating precise delineation of perirectal lymph nodes.

\subsection{Data construction}

We conducted a random split of the Meply dataset into two subsets: 214 cases for training and 55 cases for testing. As the original CT data encompassed the entire body, we took the necessary steps to enhance training efficiency by eliminating irrelevant regions. Slices not containing the rectum were removed, and the corresponding images and labels were then packed into image-label pairs. In the end, we obtained 5,624 slice pairs for training and 1,462 pairs for testing.

\section{Method}

\subsection{Overview of the CoSAM}

Inspired by the success of multi-task learning in the field of medical image processing, we constructed a collaborative learning framework for end-to-end lymph node detection and segmentation tasks based on SAM, named CoSAM. As illustrated in Figure \ref{fig:framework}, this framework encompasses a sequence-based lymph node detector, an prompt-based lymph node segmentation network, and a final collaborative processing unit that coordinates the two tasks. The detection module and the segmentation module are not in an equal parallel relationship. Leveraging the promptable paradigm of SAM, we guide the segmentation task with the spatial prior knowledge obtained from the detection module's results, ensuring consistency between segmentation and detection results, thereby ensuring the morphological integrity of the segmentation results. Moreover, our framework jointly learns detection and segmentation tasks in an end-to-end manner, where the two tasks are interdependent and mutually reinforcing.

\subsection{2.5D Sequence-based Lymph Node Detector}
Most currently available detectors face challenges in detecting perirectal lymph nodes in CT images. On the one hand, the suboptimal performance of the 2D detector can be attributed to the intrinsic three-dimensional properties of CT images and the distinctive anatomical features of perirectal lymph nodes. 
On the other hand, due to the inherent complexity of human rectal surrounding tissues and organs, the 3D detector, while introducing richer contextual information, also brings about increased background interference. 
To address this issue, we introduce a 2.5D sequence-based detector for perirectal lymph nodes detection.

To be specific, given a pre-processed CT sequence $x \in \mathbb{R}^{L \times W \times H}$, where $W$ and $H$ denote the width and height of a single CT slice, respectively, and $L$ represents the number of slices in $x$, our sequence-based detector predicts a set of bounding-boxes of suspicious perirectal lymph nodes, as well as their corresponding confidence scores.

As is shown in Figure \ref{fig:framework}, our proposed 2.5D sequence-based detector includes two stages. 
In the first stage, our method generates sequence proposals in a dense manner. Each proposal tracks frame-by-frame the possible target in a certain columnar region. 
These proposals are preliminarily screened and used for a more refined selection in the next stage.
In the second stage, the model encodes sequence features by 
integrating 2D features along the Z-axis direction under the guidance of the filtered sequence proposals. The whole process can be formulated as follows: 
\begin{equation}
\mathcal{F}_i^j = RoIP ooling(\mathcal{P}_i^j)
\end{equation}
\begin{equation}
    \mathcal{SF}_i = Concat(\mathcal{F}_i^0, \mathcal{F}_i^1, ..., \mathcal{F}_i^{L-1})
\end{equation}
where $\mathcal{P}_i^j$ denotes the $j^{th}$ 2D RoI in the $i^{th}$ sequence proposal,  $\mathcal{SF}_i$ denotes sequence features of the $i^{th}$, $L$ indicates the length of each sequence proposal.

Subsequently, in order to extract the three-dimensional spatial contextual information embedded in the sequence features, sequence features $\mathcal{SF}$ are forwarded into the transformer-based sequence processing module, which adopts a encoder-decoder framework as follows:
\begin{equation}
    \mathcal{SF'} = Encoder(\mathcal{SF})
\end{equation}
\begin{equation}
    \mathcal{F}_{det} = Decoder(\mathcal{SF'}, \mathcal{Q}_0)
\end{equation}
where the $Encoder$ and $Decoder$ indicate the transformer encoder and decoder, respectively. $\mathcal{SF'} \in \mathbb{R}^{N \times L \times D} $ denotes the sequence features encoded by transformer encoder. $\mathcal{Q}_0 \in \mathbb{R}^{N \times D}$ denotes the learnable queries, and $\mathcal{F}_{det} \in \mathbb{R}^{N \times D}$ represents the objective tokens. Eventually, the objective tokens $\mathcal{F}_{det}$  are used in the box prediction. 

\subsection{Prompt-based RoI Refinement and Segmentation}
We noticed that the SAM can not only segment specified targets based on prompts but also implicitly extract feature information of the specified RoI during this process. 
Utilizing its promptable paradigm and word lookup mechanism, we proposed a novel approach to extract morphological and anatomical information of lymph nodes, particularly their size and shape, as well as to predict their masks. 
To better extract detailed information, we adopted a variant of the SAM, namely U-SAM\cite{zhang2023care}, which incorporates a U-shaped structure and skip connections into SAM.  

Specifically, given an input CT slice $x \in \mathbb{R}^{W \times H}$ with resolution of $W \times H$ and bounding-boxes $b \in \mathbb{R}^{K \times 4}$ with a total number of K, the SAM predicts the segmentation of suspicious perirectal lymph nodes over all $K$ candidate areas. 
The generation process of partial masks and mask tokens can be formulated as follows.
\begin{equation}
    p_i = PromptEncoder(b_i)
\end{equation}
\begin{equation}
    f_x = ImageEncoder(x)
\end{equation}
\begin{equation}
    m_i, t_i = MaskDecoder(f_x, b_i)
\end{equation}
where $b_i$ denotes the $i^{th}$ bounding-box, $m_i$ denotes the predicted mask in area $b_i$, and $t_i$ represents the corresponding mask token. $PromptEncoder$, $ImageEncoder$ and $MaskDecoder$ represent the prompt encoder, the image encoder and the mask decoder of the SAM, respectively.

In the segmentation branch, all $K$ partial masks $m_i$ are collected and incorporated into the comprehensive segmentation results. In the detection branch, mask tokens $t_i$, together with sequence features $\mathcal{SF}_i$, are utilized in a joint classification head to suppress false positive results.

\section{Experiments}
\subsection{Implementation Details}
The proposed CoSAM was implemented with PyTorch 1.10. All experiments were performed on a machine with NVIDIA GTX 3090 GPUs. 
In order to enhance training stability and convergence speed, we pretrained the detector and segmentation network separately for 100 epochs. These two sub-networks were subsequently trained jointly in our collaborative learning framework for 100 epochs.  
Due to their highly dissimilar structures, distinct learning rates were employed for the detector and segmentation network. More detailed setting can be referred to the supplement materials. 
During preprocessing, CT image intensities were
truncated between [- 100, 100] Hounsfield units (HU) and then normalized to the range of [0, 1]. 
For data augmentation, we adopted random cropping, random flipping, random contrast adjustment.

\subsection{Evaluation Results}
\begin{table*}[t]
    \begin{minipage}[t]{.45\linewidth}
        \vspace{2px}
        \caption{\textbf{Results of segmentation on Meply dataset.} We compare our method with classical methods and SAM-based methods. We report Dice(\%, $\uparrow$) and IoU(\%, $\uparrow$) on the test set. }\label{tab:shape_classification}
        \label{SOTA_lymph}
        \renewcommand{\arraystretch}{1.3}
    	\begin{tabular*}{\hsize}
        {@{}@{\extracolsep{\fill}}l|c|c@{}}
    		\toprule
    		Networks & Dice & IoU\\
    		\midrule
    		U-Net\cite{ronneberger2015u}
            & 68.35 & 65.17\\
            MissFormer\cite{huang2022missformer}
            & 64.54 & 47.64\\
            TransUnet\cite{chen2021transunet}
            & 67.61 & 51.07\\
            V-Net\cite{milletari2016v}
            & 66.37 & 49.66\\
            DoubleUnet\cite{jha2020doubleu}
            & 65.40 & 48.59\\
            SwinUnet\cite{cao2022swin}
            & 68.47 & 52.05\\
            UCTransNet\cite{wang2022uctransnet}
            & 68.68 & 52.30\\
            AttenUnet\cite{oktay2018attention}
            & 65.23 & 48.40\\
            MultiResUnet\cite{ibtehaz2020multiresunet}
            & 69.05 & 52.73\\
    		\midrule
            SAM\cite{kirillov2023segment}
            & 67.74 & 46.77\\
            SAMed\cite{ma2023segment}
            & 70.79 & 54.79\\
            U-SAM\cite{zhang2023care}
            & 69.08 & 52.76\\
            \textbf{CoSAM(ours)}
            & \textbf{74.12} & \textbf{58.59}\\
    		\bottomrule
    	\end{tabular*}
    \end{minipage}
    \hspace{.1\linewidth}
    \begin{minipage}[t]{.45\linewidth}
    \centering
    
    \caption{\textbf{Results of detection module for Meply dataset.} Window size refers to the length of the input CT frame sequence, and $AP^{50}$ measures the performance of object detection.}
    \label{ablation_wsize}
    \begin{tabular*}{\hsize}
    {@{}@{\extracolsep{\fill}}c|c@{}}
        \toprule
        Window Size & $AP^{50}  \quad\quad\quad\quad$\\
        \midrule
        5 & 0.820 $\quad\quad\quad$\\
        7 & 0.835 $\quad\quad\quad$\\
        9 & 0.845 $\quad\quad\quad$\\
        11 & \textbf{0.849} $\quad\quad\quad$\\
        13 & 0.839 $\quad\quad\quad$\\
        15 & 0.810 $\quad\quad\quad$\\
        \bottomrule
    \end{tabular*}
    \vspace{5px}
    
    \caption{\textbf{Results of ablation studies on CoSAM for Meply dataset.} E2E means training end-to-end in the proposed collaborative learning framework, and CCM represents using collaborative classification module.}
    \label{ablation}
    \begin{tabular*}{\hsize}
    {@{}@{\extracolsep{\fill}}cc|c|c@{}}
        \toprule
        E2E & CCM & $AP^{50}$ & Dice\\
        \midrule
        \ding{55} & \ding{55} & 0.849 & 59.45\\
        \ding{51} & \ding{55} & 0.847 & 69.46\\
        \ding{51} & \ding{51} & \textbf{0.875} & \textbf{74.12}\\
        \bottomrule
    \end{tabular*}
    \end{minipage}
\end{table*}

\textbf{Comparisons with SOTA.}
To evaluate the effectiveness of our proposed CoSAM, we compared its segmentation performance with several state-of-the-art methods on the Meply dataset. As reported in Table \ref{SOTA_lymph}, the proposed method achieves a Dice score of 74.21\% and an IoU score of 59.00\%. 
Our method is not only superior to classical segmentation methods but also outperforms SAM-based methods. 

\textbf{Ablation Study.}
As demonstrated in Table \ref{ablation_wsize}, we conducted a sequence length ablation study on CT sequences. Experimental findings reveal that a window size of 11 yields superior performance for the detection module. Upon comprehensive consideration of model efficacy and parameter efficiency, we finalized the window size as 9, maintaining consistency across all experiments.

As shown in Table \ref{ablation}, employing an end-to-end collaborative learning framework significantly enhances the model's segmentation performance compared to independently learning the two tasks. This indicates that our proposed collaborative learning approach contributes to a more effective collaboration between the detection module and the segmentation module. Furthermore, with the addition of the collaborative classification module, our method achieves further improvements in both detection and segmentation performance. This implies that the performance enhancement of the detection module better guides the segmentation task, and conversely, the performance improvement of the segmentation module also benefits the classification accuracy of the detection task.

\textbf{Visualization Results.}

\begin{figure}[t]
    \centering
    \includegraphics[width=\linewidth,scale=0.25]{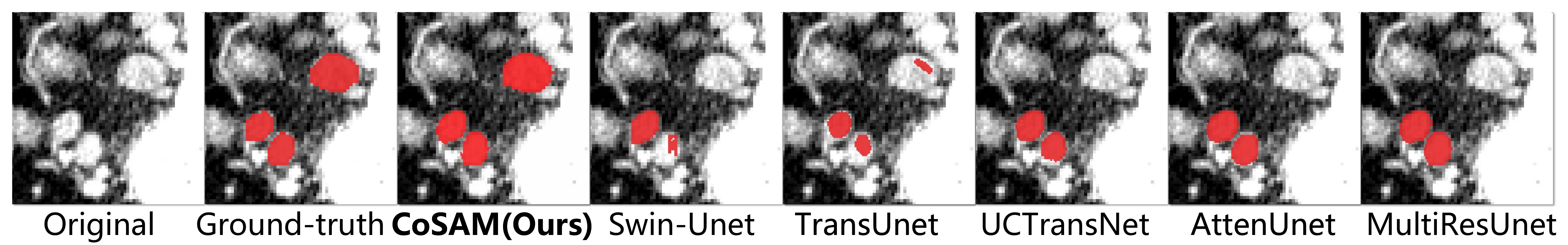}
    \caption{Visual comparisons of different segmentation methods on Meply dataset.}
    \label{fig:visualization}
\end{figure}

Figure \ref{fig:visualization} presents representative results of perirectal metastatic lymph node segmentation. It demonstrates that by establishing strong consistency between detection and segmentation, our method can better ensure morphological integrity and prevent false positive segmentation.

\section{Conclusion}
In this paper, we present Meply, the first large-scale, finely annotated dataset for segmenting metastatic lymph nodes in the context of rectal cancer. Additionally, for the task of segmenting metastatic lymph nodes around the rectum, we apply the Segment Anything Model (SAM)\cite{kirillov2023segment} prompt mechanism to medical segmentation, proposing a CoSAM framework based on SAM for collaborative learning of rectal perirectal lymph node detection and segmentation. We conduct a series of experiments on the Meply dataset to validate its effectiveness.

\textbf{Acknowledgement.} This work is supported by The University Synergy Innovation Program of Anhui Province (Grant No. GXXT-2022-056).
\bibliographystyle{splncs04}
\bibliography{mybibliography}
\end{document}